\documentclass{article}
\usepackage{spconf,amsmath,graphicx}
\usepackage{multirow}

\usepackage[utf8]{inputenc}
\usepackage{microtype}

\title{Capitalization Normalization for Language Modeling with an Accurate and Efficient Hierarchical RNN Model}
\name{Hao Zhang, You-Chi Cheng, Shankar Kumar, W. Ronny Huang, Mingqing Chen, Rajiv Mathews}
\address{Google Research\\
\texttt{haozhang@google.com}
}
\begin{document}
%
\maketitle
\begin{abstract}
Capitalization normalization (truecasing) is the task of restoring the correct case (uppercase or lowercase) of noisy text. 
We propose a fast, accurate and compact two-level hierarchical word-and-character-based recurrent neural network model. We use the truecaser to normalize user-generated text in a Federated Learning framework for language modeling. A case-aware language model trained on this normalized text achieves the same perplexity as a model trained on text with gold capitalization. In a real user A/B experiment, we demonstrate that the improvement translates to reduced prediction error rates in a virtual keyboard application. Similarly, in an ASR language model fusion experiment, we show reduction in uppercase character error rate and word error rate.
\end{abstract}
\begin{keywords}
Text normalization, capitalization, truecasing, language modeling.
\end{keywords}
\section{Introduction}
\label{sec:intro}
The vast amount of online text powers language models for speech recognition, typing suggestions and many other language generation tasks. However user-generated texts, especially those from mobile applications such as Twitter Tweets \cite{DBLP:conf/www/NebhiBG15}, often violate the grammatical rules of casing in English and other western languages \cite{grishina-etal-2020-truecasing}.  The process of restoring the proper case, often known as \textbf{t{R}u{E}cas{I}ng} \cite{lita-etal-2003-truecasing}, provides a factorized solution with a dedicated model for case normalization.
Such a model can be applied on noisy text before case-aware language model training or on model output after case-agnostic language model inference.
Concretely, it has been used for restoring case and improving the recognition performance for ASR ({\em post-processing}, \cite{4960690, 6638968, sunkara-etal-2020-robust}) as well as case normalization of user text prior to language model (LM) training ({\em pre-processing}, \cite{DBLP:conf/conll/ChenSMWABR19}). \cite{DBLP:conf/conll/ChenSMWABR19} employs Federated Learning \cite{DBLP:conf/aistats/McMahanMRHA17}, a privacy-preserving learning paradigm, on distributed devices. The need for normalizing text on a variety of mobile devices makes it an imperative to develop a fast, accurate and compact truecaser.

From a modeling perspective, prior work can be grouped into  word-based and character-based approaches.
Most of the earlier works are word-based \cite{lita-etal-2003-truecasing, chelba-acero-2004-adaptation}.
In the word-based view, the task is to classify each word in the input into one of the few classes \cite{lita-etal-2003-truecasing}: all lowercase (\texttt{LC}), first letter uppercase (\texttt{UC}), all letters uppercase (\texttt{CA}), and mixed case (\texttt{MC}). The main drawback of the word-based approach is that it does not generalize well to unseen words and mixed case words.
The development of character-based neural network modeling techniques \cite{pmlr-v32-santos14} led to the introduction of character-based truecasers \cite{susanto-etal-2016-learning, 9031494}.
In the character-based view, the task is to classify each character in the input into one of the two classes: \texttt{U} and \texttt{L}, for uppercase and lowercase respectively.
The shortcoming of character-based models is inefficiency. As word-level features in some form are important for classification, character-based models need to be deep enough to capture word-level features, which exacerbates their slowness. 

We view truecasing as a special case of text normalization \cite{DBLP:journals/coling/ZhangSNSPGR19} with similar trade-offs for accuracy and efficiency. Thus, we classify input words into one of two classes: \texttt{SELF} and \texttt{OTHER}. Words labeled \texttt{SELF} will be copied as-is to the output. Words labeled \texttt{OTHER} will be fed to a sub-level character-based neural network along with the surrounding context, which is encoded bidirectionally. The task of the sub-network is to perform a non-trivial transduction such as:
\texttt{iphone} $\rightarrow$ \texttt{iPhone},
\texttt{mcdonald's} $\rightarrow$ \texttt{McDonald's} 
and \texttt{hewlett-packard} $\rightarrow$  \texttt{Hewlett-Packard}.

The two-level hierarchical network is fast while also being accurate. We report the model's intrinsic accuracy, speed, and size on a curated Wikipedia data set in Section~\ref{sec:large_scale}. Next, we focus on case-aware language modeling on noisy text. 
In Section~\ref{sec:lm_ppl}, we show that applying the model in preprocessing can reduce a cased language model's perplexity. Furthermore, in Section~\ref{sec:lm_keyboard}, we show that when the cased language model trained in this fashion is applied in a noisy channel virtual keyboard, it leads to reduction in word prediction errors in an A/B test on real users. Finally, in section~\ref{sec:lm_asr}, we show similar improvement in ASR language model fusion.

\section{Related Work}
\label{sec:related}

Word-based truecasing has been the dominant approach since the introduction of the task by \cite{lita-etal-2003-truecasing}.
Word-based models can be further categorized into generative models such as HMMs \cite{lita-etal-2003-truecasing, 4960690, 6638968, DBLP:conf/www/NebhiBG15} and
discriminative models such as Maximum-Entropy Markov Models \cite{chelba-acero-2004-adaptation}, Conditional Random Fields \cite{DBLP:conf/naacl/WangKM06}, and most recently Transformer neural network models \cite{9041202, 10.1007/978-3-030-50146-4_52, sunkara-etal-2020-robust}.
Word-based models need to refine the class of mixed case words because there is a  combinatorial number of possibilities of case mixing for a word (e.g., \texttt{LaTeX}). \cite{lita-etal-2003-truecasing, chelba-acero-2004-adaptation} suggested using either all of the forms or the most frequent form of mixed case words in the training data. The large-scale finite state transducer (FST) models in \cite{6638968} used all known forms of mixed case words to build the ``capitalization'' FST. \cite{DBLP:conf/naacl/WangKM06} used a heuristic \textbf{GEN} function to enumerate a superset of all forms seen in the training data.
But others \cite{9041202, 10.1007/978-3-030-50146-4_52} have chosen to simplify the problem by mapping mixed case words to first letter uppercase words. \cite{sunkara-etal-2020-robust} only evaluated word class F1, without refining the class of \texttt{MC}.

Character-based models have been explored largely after the dawn of modern neural network models. \cite{susanto-etal-2016-learning} first introduced character-based LSTM for this task and completely solved the mixed case word problem. Recently, \cite{grishina-etal-2020-truecasing} compared character-based $n$-gram ($n$ up to 15) language models with the character LSTM of \cite{susanto-etal-2016-learning}.
\cite{9031494} advanced the state of the art with a character-based CNN-LSTM-CRF model.

Text normalization is the process of transforming text into a canonical form. Examples of text normalization include but are not limited to written-to-spoken text normalization for speech synthesis \cite{DBLP:journals/coling/ZhangSNSPGR19}, spoken-to-written text normalization for speech recognition \cite{Peyser2019}, social media text normalization \cite{10.1145/2414425.2414430}, and historical text normalization \cite{makarov-clematide-2020-semi}. Truecasing is a problem that appears in both spoken-to-written and social media text normalization.

\section{Formulation and Model Architecture}
\label{sec:problem_and_model}
The input is a sequence of all lowercase words $\vec{\mathrm{X}}=(x_1,\dots,x_l)$. The output is a sequence of words with proper casing $\vec{\mathrm{Y}}=(y_1,\dots,y_l)$. We introduce a latent sequence of class labels $\vec{\mathrm{C}}=(c_1,\dots,c_l)$, where $c_i \in \{\texttt{S=SELF,O=OTHER}\}$. We use the notation $x_i^j$ and $y_i^j$ to represent the $j\text{-th}$ character within the $i\text{-th}$ word.

The model is trained to predict the probability: 
\begin{equation}
\label{eqn:hier}
    \mathrm{P}(\vec{\mathrm{Y}}|\vec{\mathrm{X}}) = \sum_{\vec{\mathrm{C}}}\mathrm{P}(\vec{\mathrm{Y}}|\vec{\mathrm{X}}, \vec{\mathrm{C}}) \cdot \mathrm{P}(\vec{\mathrm{C}}|\vec{\mathrm{X}}),
\end{equation}
where $\mathrm{P}(\vec{\mathrm{C}}|\vec{\mathrm{X}})$ is a word-level model that predicts if a word should stay all lowercase (\texttt{SELF}) or change to a different case form (\texttt{OTHER}) taking into account label dependencies between $c_1,\dots,c_{i-1}$ and $c_i$.
\begin{equation}
\label{eqn:word_based}
    \mathrm{P}(\vec{\mathrm{C}}|\vec{\mathrm{X}})  =  \prod_{i=1}^{l} {\mathrm{P}(c_i|c_1,\dots,c_{i-1},\vec{\mathrm{X}}})
\end{equation}
The label sequence $\vec{\mathrm{C}}$ works as a gating mechanism,
\begin{align}
\label{eqn:dispatch}
    \mathrm{P}(\vec{\mathrm{Y}}|\vec{\mathrm{X}}, \vec{\mathrm{C}}) =& \prod_{i=1}^{l}{\delta(c_i, \texttt{O})  \mathrm{P}(y_i|{\mathrm{X}) + \delta(c_i, \texttt{S})} \delta(x_i, y_i)},
\end{align}
where $\mathrm{P}(y_i|{\mathrm{X}})$ is a character-level model that predicts each output character within a word, assuming dependency between characters within a word: $y_i^{1},\dots,y_i^{j-1}$ and $y_i^j$, but no cross-word dependency between $y_1,\dots,y_{i-1}$ and $y_i$, and \texttt{S} and \texttt{O} denote \texttt{SELF} and \texttt{OTHER} respectively.
\begin{equation}
\label{eqn:char_based}
    \mathrm{P}(y_i|{\mathrm{X}}) = \prod_{j=1}^{j=|x_i|}{\mathrm{P}(y_i^j|y_i^{1},\dots,y_i^{j-1},\vec{\mathrm{X}})}
\end{equation}
Given that $\delta(c_i, \texttt{S}) \equiv \delta(x_i, y_i)$, 
we can derive the log likelihood of Equation~\ref{eqn:hier} as:
\begin{align}
    \log(\mathrm{P}(\vec{\mathrm{Y}}|\vec{\mathrm{X}})) &= \sum_{i=1}^{l}{\delta(c_i, \texttt{O}) \log(\mathrm{P}(y_i|{\mathrm{X}}))} \nonumber \\
    &+\log(\mathrm{P}(\vec{\mathrm{C}}|\vec{\mathrm{X}}))
\end{align}
Equations~\ref{eqn:word_based} and ~\ref{eqn:char_based} can be modeled as sequence-to-sequence (seq2seq) problems. Unlike the general machine translation problem with unequal number of input and output tokens which requires a soft attention mechanism \cite{DBLP:journals/corr/BahdanauCB14}, both of our seq2seq problems can assume hard alignment between the output label at each time step and the input symbol at the same time step ($c_i$ is aligned to $x_i$, $y_i^j$ is aligned to $x_i^j$).

\begin{figure}[hbt!]
  \includegraphics[width=1.2\columnwidth]{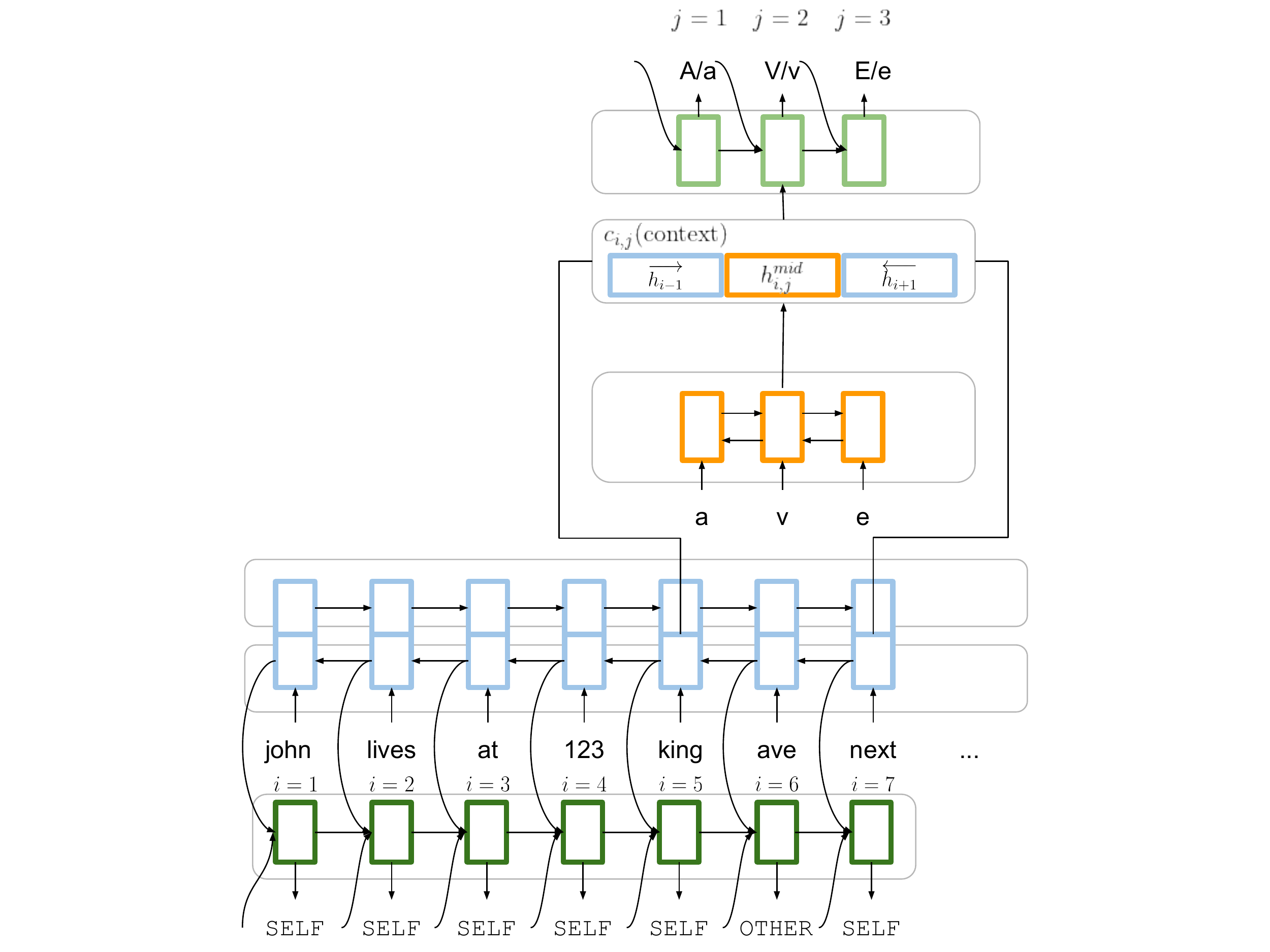}
\caption{\label{fig:architecture} Hierarchical RNN architecture.} 
\end{figure}

We follow the multi-task recurrent neural network architecture of \cite{DBLP:journals/coling/ZhangSNSPGR19}.
Figure~\ref{fig:architecture} displays the rolled-out network of the two-level hierarchical RNN on a typical input sentence. The key difference with \cite{DBLP:journals/coling/ZhangSNSPGR19} is that in our setting, the models at both levels perform sequence tagging whereas their second-level model is a full-fledged sequence-to-sequence model with soft attention \cite{DBLP:journals/corr/BahdanauCB14}.

The exact inference requires searching over all possible $\vec{\mathrm{C}}$ and  $\vec{\mathrm{Y}}$ which is infeasible for RNNs. But we can do beam search for both Equation~\ref{eqn:word_based} and Equation~\ref{eqn:char_based} and use either the entire beam of hypotheses of $\vec{\mathrm{C}}$ or simply the best one in Equation~\ref{eqn:dispatch}. In our experiments, using the entire beam only helped slightly.

\section{Experiments}
\label{sec:expts}
Our models as depicted in Figure~\ref{fig:architecture} have a pair of forward and backward RNNs over words to encode sentence context. 
Words are first encoded based on subword features, namely character $n$-grams with $n$ up to 3 \cite{DBLP:journals/coling/ZhangSNSPGR19}. In the example shown in Figure~\ref{fig:architecture}, the character $n$-grams representing the word \texttt{ave} include \texttt{a}, \texttt{v}, \texttt{e}, \texttt{<s>a}, \texttt{av}, \texttt{ve}, \texttt{e<s>}, \texttt{<s>av}, \texttt{ave}, and \texttt{ve<s>}, where \texttt{<s>} is the placeholder symbol for the left and right boundaries.
Each word is represented as the summation of the embeddings of the hashed character $n$-grams.

We train a large teacher model and then apply sequence distillation \cite{kim-rush-2016-sequence} to produce a compact student model.
The hyper-parameters of both models are listed in Table~\ref{tab:network_specs}. We use the same hyper-parameters for the word- and character-level sub-models.

\begin{table}[ht]
    \centering
    \begin{small}
    \begin{tabular}{lrr}
                                & teacher & student \\                               
                                & (large) & (small) \\
             \hline
             input embedding size & 512 & 128 \\
             output embedding size & 512 & 128 \\
            \# of forward enc. layers & 2 & 1\\
             \# of backward enc. layers & 2 & 1\\
             \# of dec. layers & 2 & 1\\
             \# of enc. GRU cells & 512 & 128\\
             \# of dec. GRU cells & 512 & 128 \\
             max char $n$-gram order & 3 & 3\\
             buckets of char $n$-grams & 5000 & 5000 \\
             beam size & 2 & 2\\
             \hline
    \end{tabular}
    \caption{Network hyper-parameters.}
        \label{tab:network_specs}  
    \end{small}
\end{table}

\subsection{Accuracy, Speed, and Model Size Comparison}
\label{sec:large_scale}
\begin{table*}[htb]
    \centering
    \begin{tabular}{rl|c|c|c|r|r}
    \multicolumn{2}{c}{{\em System}} & {\em Precision} & {\em Recall} & {\em F1} & {\em Speed} & {\em \# of params}\\
    \hline
    \multirow{2}{*}{5-gram FST} & 
    unpruned, unoptimized & 91.64 & 43.55 & 59.04 & 1.0x & 10M\\
    &pruned, optimized& \textbf{91.88} & 41.19 & 56.88 & 88.0x & 1M \\
    \hline
    \multirow{4}{*}{char. RNN}
    &small, 1-layer uni-, dec. & 69.11 & 22.86 & 34.35 & 0.7x & 230K \\
    &small, 1-layer bi-, enc.\&dec. & 86.12 & 75.07 & 80.22 & 0.5x & 400K\\
    &large, 2-layer bi-, enc.\&dec. & 87.06 & 78.09 & 82.33 & 0.1x & 8.4M\\
    \hline
    \multirow{1}{*}{hier. RNN, student} & small, 1-layer bi-, enc.\&dec. $\times$2 & 86.95  & 79.81 & 83.23 & 2.2x & 1.3M\\
    \multirow{1}{*}{hier. RNN, teacher} & large, 2-layer bi-, enc.\&dec. $\times$2 & 88.01  & \textbf{82.60} & \textbf{85.22} & 0.3x & 19.2M \\
    \end{tabular}
    \caption{Large-scale comparison across model types on intrinsically capitalized Wikipedia edit history data set.}
    \label{tab:wiki_edit}
\end{table*}


In this section, we report performance of models trained on a large data set of digitized books and newswire articles containing 1.5 billion sentences. We use 90\% of the data for training and sample a subset from the remaining 10\% for validation. We use a human-curated data set of 1200 sentences\footnote{https://github.com/google-research-datasets/wikipedia-intrinsic-capitalization} from Wikipedia edit history \cite{lichtarge-etal-2019-corpora} for testing. 

All models take lowercase tokenized sentences as input and output the same tokens with predicted casing. The predicted tokens are compared against references.
Following \cite{susanto-etal-2016-learning}, we use non-lowercase (NL) F1 as the evaluation metric.
\begin{small}
\begin{align*}
\text{NL Precision} = &\frac{\text{\# of correct NL predictions}}{\text{\# of NL predictions}} \\
\text{NL Recall} = &\frac{\text{\# of correct NL predictions}}{\text{\# of NL references}}
\end{align*}
\end{small}
NL F1 is the harmonic mean of NL Precision and Recall. For brevity, we use precision, recall and F1 without qualification.

We present three groups of results in Table~\ref{tab:wiki_edit}. The FST group represents word-based HMM models with a wide coverage of mixed case words. The character-based RNN group is our implementation of  \cite{susanto-etal-2016-learning} with various hyper-parameters. The third group is our hierarchical RNNs.


We report CPU speed relative to the large FST model. We use batch size of 1 at inference time for all models. All models are implemented in TensorFlow and quantized after training. We compare the model sizes in terms of the total number of parameters. 

FST models have high precision but low recall, showing the coverage problem of word-based models. The one-layer uni-directional character-based RNN scores low in both precision and recall. 
Even a single-layer bidirectional character-based RNN is already 4 times slower than a single-layer word- and character-based hierarchical RNN. Using two layers in both the encoder and the decoder, a purely character-based model is nearly 20 times slower than a hierarchical model with about the same F1 score. 

\subsection{Case-aware Language Models}
\label{sec:lm_ppl}
In this section, we study the effect of capitalization for case-aware language models. We use the 1B Word Benchmark data set (LM1B) introduced by ~\cite{chelba2013lm1b}. The training and evaluation sections consist of 0.8B words (30.4M sentences) and 153K words (6075 sentences) respectively. The vocabulary consists of 800K words.  To simulate the scenario when the training data has noisy capitalization, we noisify the training data by randomly lowercasing either 25\% or 50\% of the capitalized words. The baseline system does not perform any capitalization normalization. Two systems to contrast with the baseline apply the FST capitalizer and the RNN capitalizer respectively. Without loss of generality, we use a moderately large (2-layer) sub-word (16k word pieces, 2048 hidden units) LSTM LM as the language model architecture~\cite{jozefowicz2016lstm}. 

Table~\ref{tab:pplx} shows the word-level perplexities of the baseline (noisy capitalization), the two capitalization normalizers (predicted capitalization), and the oracle (ground-truth capitalization). We observe that the RNN capitalizer yields a similar perplexity as the oracle while outperforming the FST capitalizer by a small margin.

\begin{table}[htb]
    \centering
    \begin{tabular}{r|c}
           {\em Capitalization Model} & {\em Perplexity} \\
         \hline
           50\% corrupt & 59.41 \\
           25\% corrupt & 54.68 \\ 
         \hline
          5-gram FST & 51.74 \\
          hier. RNN & \textbf{51.60} \\
         \hline
          oracle & 51.61\\
    \end{tabular}
    \caption{Perplexities of RNN language models on LM1B using different capitalization normalization methods.}
    \label{tab:pplx}
\end{table}

\subsection{Case-aware Language Models in Virtual Keyboard Applications}
\label{sec:lm_keyboard}
Virtual keyboards are indispensable for text input on the popular touchscreen devices. As spatial input signals of tap and glide trails are often noisy, decoding the intended input can be achieved with a noisy channel model composed of a spatial model and a language model \cite{ouyang2017mobile}. The quality of language models directly impacts the prediction accuracy of virtual keyboards which is reflected in features such as auto-corrections, word completions, and next word predictions. In this section, we report the results of an A/B experiment on real users. The two systems under comparison differ only in the cased language models they use in decoding. The baseline uses the pruned FST model for capitalization normalization. The new system uses the hierarchical RNN model for capitalization normalization. The normalization step is done in a distributed fashion on users' devices before privacy-preserving Federated Learning of neural language models \cite{DBLP:conf/conll/ChenSMWABR19} starts. The two resulting LMs have exactly the same architecture \cite{DBLP:conf/conll/ChenSMWABR19} and the same number of parameters. Both are trained for the same number of rounds which amount to billions of tokens.

Table~\ref{tab:a_b_results} shows the results of the A/B test using metrics indicative of the prediction accuracy of the virtual keyboard. The test covers a time span long enough to accumulate statistics over more than 1 billion typed words. We observe reductions of word modification rate (WMR) and auto-correction rejection rate (RAC) in the virtual keyboard usage statistics. Even though references are impossible to obtain in the A/B test, these metrics approximate WER (Word Error Rate) with regard to the intended user input.
The reductions are statistically significant as shown by the 95\% confidence intervals. We also observe a small but non-significant reduction of Out-of-vocabulary (OOV) rate by 0.4\%. The OOV rate changes because the vocabulary size is identical in both LMs. Capitalization normalization causes the frequency ranking of word types to change.

\begin{table}[htb]
    \centering 
    \begin{tabular}{c|c|c}
        {\em Model} & {\em WMR} & {\em RAC} \\
         \hline
         5-gram FST & 5.81\% & 2.91\%\\
         hier. RNN & 5.78\%  & 2.87\%\\ \hline
         Rel. Reduction & \textbf{[-0.92,-0.11]\%}& \textbf{[-2.21, -0.69]\%}\\
    \end{tabular}
    \caption{Virtual keyboard A/B experiment results. {\em WMR} is the fraction of words modified or retyped. {\em RAC} is the auto-correction rejection rate. The last row shows the 95\% confidence interval of the relative reductions.}
    \label{tab:a_b_results}
\end{table}

\subsection{Case-aware Language Models in Speech Recognition}
\label{sec:lm_asr}
Our speech recognizer is a streaming Conformer acoustic model \cite{sainath2021} integrated with a Conformer language model \cite{gulati2020conformer} via HAT shallow fusion \cite{variani2020}, and we measure performance on a $\sim$10k sample of Google voice search traffic with natively capitalized reference transcripts. When the acoustic model is fixed and the output tokens are cased, we show that improving capitalization normalization of the language model training data leads to reduction of upper-case error rate (UER), the character error rate of each capitalized character in either the predicted or reference transcript (Table~\ref{table:asr_results}). 
Importantly, the overall case-insensitive WER is also reduced, likely due to the language model inducing more favorable beam search decisions.

\begin{table}[htb]
   \centering 
   \begin{tabular}{c|c|c}
           {\em Model} & {\em WER} & {\em UER} \\
         \hline
         5-gram FST & 5.8 & 32.6 \\
         hier. RNN & \textbf{5.6} & \textbf{32.4} \\
   \end{tabular}
   \caption{ASR LM fusion experiment results. The two systems in comparison differ only in the capitalization normalization model used to pre-process the LM training data.}
   \label{table:asr_results}
\end{table}

\section{Conclusions}
Truecasing provides a factored solution to improve case-aware language modeling for applications such as ASR and 
text input in virtual keyboards. We propose a hierarchical word-and-character-based RNN model with the speed advantage of word-based models and accuracy advantage of character-based models. The model is efficient enough to be uploaded to mobile devices to train a language model using Federated Learning. The improvement is manifested in reduction of prediction error rates in a large-scale A/B experiment using a virtual keyboard and an ASR LM fusion experiment.
\bibliographystyle{IEEEbib}
\bibliography{custom}

\begin{thebibliography}{10}

\bibitem{DBLP:conf/www/NebhiBG15}
K. Nebhi, K. Bontcheva, and G. Gorrell,
\newblock ``Restoring capitalization in \#tweets,''
\newblock in {\em Proc. WWW}, 2015.

\bibitem{grishina-etal-2020-truecasing}
Y. Grishina, T. Gueudre, and R. Winkler,
\newblock ``Truecasing {G}erman user-generated conversational text,''
\newblock in {\em Proc. W-NUT}, 2020.

\bibitem{lita-etal-2003-truecasing}
L.~V. Lita, A. Ittycheriah, S. Roukos, and N. Kambhatla,
\newblock ``t{R}u{E}cas{I}ng,''
\newblock in {\em Proc. ACL}, 2003.

\bibitem{4960690}
A. Gravano, M. Jansche, and M. Bacchiani,
\newblock ``Restoring punctuation and capitalization in transcribed speech,''
\newblock in {\em Proc. ICASSP}, 2009.

\bibitem{6638968}
F. Beaufays and B. Strope,
\newblock ``Language model capitalization,''
\newblock in {\em Proc. ICASSP}, 2013.

\bibitem{sunkara-etal-2020-robust}
M. Sunkara, S. Ronanki, K. Dixit, S. Bodapati, and K. Kirchhoff,
\newblock ``Robust prediction of punctuation and truecasing for medical
  {ASR},''
\newblock in {\em Proc. NLPMC}, 2020.

\bibitem{DBLP:conf/conll/ChenSMWABR19}
M. Chen, A.~T. Suresh, R. Mathews, A. Wong, C. Allauzen, F. Beaufays, and M.
  Riley,
\newblock ``Federated learning of n-gram language models,''
\newblock in {\em Proc. CoNLL}, 2019.

\bibitem{DBLP:conf/aistats/McMahanMRHA17}
B. McMahan, E. Moore, D. Ramage, S. Hampson, and B.~A. y~Arcas,
\newblock ``Communication-efficient learning of deep networks from
  decentralized data,''
\newblock in {\em Proc. AISTATS}, 2017.

\bibitem{chelba-acero-2004-adaptation}
C. Chelba and A. Acero,
\newblock ``Adaptation of maximum entropy capitalizer: Little data can help a
  lot,''
\newblock in {\em Proc. EMNLP}, 2004.

\bibitem{pmlr-v32-santos14}
C.~D. Santos and B. Zadrozny,
\newblock ``Learning character-level representations for pos tagging,''
\newblock in {\em Proc. ICML}, 2014.

\bibitem{susanto-etal-2016-learning}
R.~H. Susanto, H.~L. Chieu, and W. Lu,
\newblock ``Learning to capitalize with character-level recurrent neural
  networks: An empirical study,''
\newblock in {\em Proc. EMMLP}, 2016.

\bibitem{9031494}
G. Ramena, D. Nagaraju, S. Moharana, D. Prasanna~Mohanty, and N. Purre,
\newblock ``An efficient architecture for predicting the case of characters
  using sequence models,''
\newblock in {\em Proc. ICSC}, 2020.

\bibitem{DBLP:journals/coling/ZhangSNSPGR19}
H. Zhang, R. Sproat, A.~H. Ng, F. Stahlberg, X. Peng, K. Gorman, and B. Roark,
\newblock ``Neural models of text normalization for speech applications,''
\newblock {\em Comput. Linguistics}, vol. 45, no. 2, 2019.

\bibitem{DBLP:conf/naacl/WangKM06}
W. Wang, K. Knight, and D. Marcu,
\newblock ``Capitalizing machine translation,''
\newblock in {\em Proc. HLT/NAACL}, 2006.

\bibitem{9041202}
B. Nguyen, V.~B.~H. Nguyen, H. Nguyen, P.~N. Phuong, T.-L. Nguyen, Q.~T. Do,
  and L.~C. Mai,
\newblock ``Fast and accurate capitalization and punctuation for automatic
  speech recognition using transformer and chunk merging,''
\newblock in {\em Proc. O-COCOSDA}, 2019.

\bibitem{10.1007/978-3-030-50146-4_52}
R. Rei, N.~M. Guerreiro, and F. Batista,
\newblock ``Automatic truecasing of video subtitles using bert: A multilingual
  adaptable approach,''
\newblock in {\em Info. Proc. and Mgmt. of Uncertainty in Knowledge-Based
  Syst.}, 2020.

\bibitem{Peyser2019}
C. Peyser, H. Zhang, T.~N. Sainath, and Z. Wu,
\newblock ``{Improving Performance of End-to-End ASR on Numeric Sequences},''
\newblock in {\em Proc. Interspeech}, 2019.

\bibitem{10.1145/2414425.2414430}
B. Han, P. Cook, and T. Baldwin,
\newblock ``Lexical normalization for social media text,''
\newblock {\em ACM Trans. Intell. Syst. Technol.}, vol. 4, no. 1, 2013.

\bibitem{makarov-clematide-2020-semi}
P. Makarov and S. Clematide,
\newblock ``Semi-supervised contextual historical text normalization,''
\newblock in {\em Proc. ACL}, 2020.

\bibitem{DBLP:journals/corr/BahdanauCB14}
D. Bahdanau, K. Cho, and Y. Bengio,
\newblock ``Neural machine translation by jointly learning to align and
  translate,''
\newblock in {\em Proc. ICLR}, 2015.

\bibitem{kim-rush-2016-sequence}
Y. Kim and A.~M. Rush,
\newblock ``Sequence-level knowledge distillation,''
\newblock in {\em Proc. EMNLP}, 2016.

\bibitem{lichtarge-etal-2019-corpora}
J. Lichtarge, C. Alberti, S. Kumar, N. Shazeer, N. Parmar, and S. Tong,
\newblock ``Corpora generation for grammatical error correction,''
\newblock in {\em Proc. NAACL}, 2019.

\bibitem{chelba2013lm1b}
C. Chelba, T. Mikolov, M. Schuster, Q. Ge, T. Brants, and P. Koehn,
\newblock ``One billion word benchmark for measuring progress in statistical
  language modeling,''
\newblock {\em CoRR}, vol. abs/1312.3005, 2013.

\bibitem{jozefowicz2016lstm}
R. J{\'{o}}zefowicz, O. Vinyals, M. Schuster, N. Shazeer, and Y. Wu,
\newblock ``Exploring the limits of language modeling,''
\newblock {\em CoRR}, vol. abs/1602.02410, 2016.

\bibitem{ouyang2017mobile}
T. Ouyang, D. Rybach, F. Beaufays, and M. Riley,
\newblock ``Mobile keyboard input decoding with finite-state transducers,''
\newblock {\em arXiv:1704.03987}, 2017.

\bibitem{sainath2021}
T.~N. Sainath, Y.~R. He, A. Narayanan, R. Botros, R. Pang, D.~J. Rybach, C.
  Allauzen, E. Variani, J. Qin, Q.-N. Le-The, A. Gruenstein, A. Gulati, B. Li,
  C. Peyser, C.-C. Chiu, D.~A. Caseiro, E. Guzman, I.~C. McGraw, J. Yu, M.~D.
  Riley, P. Rondon, Q. Liang, S. Mavandadi, S. yiin Chang, T.~D. Strohman,
  W.~R. Huang, W. Li, Y. Wu, and Y. Zhang,
\newblock ``An efficient streaming non-recurrent on-device end-to-end model
  with improvements to rare-word modeling,''
\newblock in {\em Proc. Interspeech}, 2021.

\bibitem{gulati2020conformer}
A. Gulati, J. Qin, C.-C. Chiu, N. Parmar, Y. Zhang, J. Yu, W. Han, S. Wang, Z.
  Zhang, Y. Wu, and R. Pang,
\newblock ``Conformer: Convolution-augmented transformer for speech
  recognition,''
\newblock in {\em Proc. Interspeech}, 2020.

\bibitem{variani2020}
E. Variani, D. Rybach, C. Allauzen, and M. Riley,
\newblock ``Hybrid autoregressive transducer (hat),''
\newblock in {\em Proc. ICASSP}, 2020.

\end{thebibliography}

\end{document}